\documentclass[letterpaper, 10 pt, conference]{ieeeconf}  

\IEEEoverridecommandlockouts                              

\usepackage{times}
\usepackage{epsfig}
\usepackage{graphicx}
\usepackage{amsmath}
\usepackage{amssymb}
\usepackage{verbatim}
\usepackage{mathtools}
\let\labelindent\relax
\usepackage{enumitem}
\usepackage{booktabs}
\usepackage{xcolor}
\usepackage{multirow}
\usepackage{subcaption}
\usepackage{algorithm} 
\usepackage{algpseudocode} 
\usepackage[pagebackref=true,breaklinks=true,letterpaper=true,colorlinks,citecolor=blue,linkcolor=blue,bookmarks=false]{hyperref}

\usepackage{pifont}
\newcommand{\xmark}{\ding{55}}%

\title{\LARGE \bf
RiskBench: A Scenario-based Benchmark for Risk Identification
}

\author{Chi-Hsi Kung$^{1}$ \, Chieh-Chi Yang$^{1}$ \, Pang-Yuan Pao$^{1}$ \, Shu-Wei Lu$^{1}$ \,  Pin-Lun Chen$^{1}$  \\ Hsin-Cheng Lu$^{2}$ \, Yi-Ting Chen$^{1\ddagger}$ \\
\thanks{$^{1.}$ C.-H. Kung, C.-C. Yang, S.-W. Lu, P.-Y. Pao, P.-L. Chen, and Y.-T. Chen is with the Department of Computer Science, National Yang Ming Chiao Tung University, Hsinchu, Taiwan.}
\thanks{$^{2.}$ H.-C. Lu is with the Department of Computer Science, National Taiwan University, Taipei, Taiwan. 
}
\thanks{$^{\ddagger}$ Corresponding Author.}
}

\begin{document}

\maketitle
\thispagestyle{empty}
\pagestyle{empty}

\begin{abstract}
Intelligent driving systems aim to achieve a zero-collision mobility experience, requiring interdisciplinary efforts to enhance safety performance. This work focuses on risk identification, the process of identifying and analyzing risks stemming from dynamic traffic participants and unexpected events. While significant advances have been made in the community, the current evaluation of different risk identification algorithms uses independent datasets, leading to difficulty in direct comparison and hindering collective progress toward safety performance enhancement. To address this limitation, we introduce \textbf{RiskBench}, a large-scale scenario-based benchmark for risk identification. We design a scenario taxonomy and augmentation pipeline to enable a systematic collection of ground truth risks under diverse scenarios. We assess the ability of ten algorithms to (1) detect and locate risks, (2) anticipate risks, and (3) facilitate decision-making. We conduct extensive experiments and summarize future research on risk identification. Our aim is to encourage collaborative endeavors in achieving a society with zero collisions. We have made our dataset and benchmark toolkit publicly at this \href{https://hcis-lab.github.io/RiskBench/}{project webpage}.
\end{abstract}

\section{Introduction}

\label{sec:intro}

%
\begin{figure}[t!]
\centering
\includegraphics[width=1.0\columnwidth,clip]{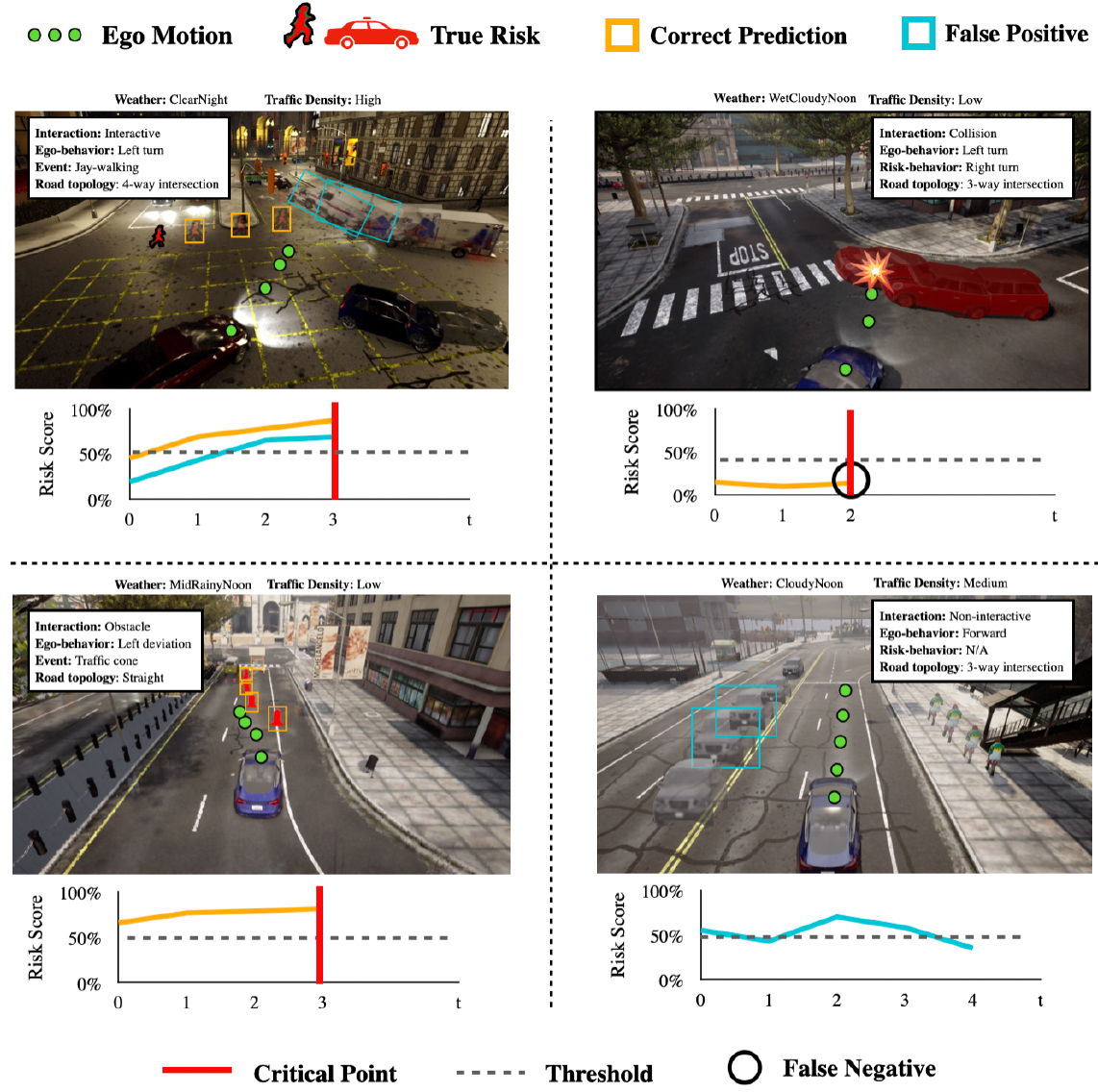} 
\caption{\textbf{Scenario-based Evaluation for Risk Identification.} We establish four distinct interaction types to cover various risk definitions explored in the community. RiskBench evaluates an algorithm's ability to identify risks stemming from dynamic traffic participants and unexpected events based on localization, anticipation, and planning awareness.}
\vspace{-.6cm}
\label{risk}

\end{figure}

In pursuit of a collision-free mobility society, numerous industries and academic institutions have dedicated their efforts to enhancing the safety capabilities of Intelligent Driving Systems (IDS).
This paper primarily centers on risk identification, the process of identifying events that could influence the decision-making of IDS.
Specifically, we delve into the evaluation of algorithms' ability to identify and assess risks stemming from dynamic traffic participants and unexpected events, such as collisions and construction zones. 
Predicting potential hazards is an inherently challenging aspect of risk identification.
%
A variety of risk identification algorithms have been attempted, such as collision anticipation~\cite{chan2016anticipating,suzuki2018anticipating,Herzig_iccvw2019,fang2019dada,You_eccv2020},
%
%
%
trajectory prediction and collision checking~\cite{Minderhoudttc2001,Althoff_srs_iv2008,lawitzky2013interactive,risk_assessment_Lefevre_ROBOMECH,Trajectron_iccv2019,Koschi_set_TIV2020,TraPHic_CVPR2019,Trajectron++_eccv2020,titan_CVPR2020,Neumann_CVPR2021},
object importance~\cite{Spain_importance_eccv2008,Ohn-Bar_object_importance_2017,Zeng_risk_cvpr2017,Gao_goal_icra2019,Zhang_graph_important_icra2020},
human gaze prediction~\cite{Alletto_Dreye_cvprw2016,Xia_ACCV_2018,Xia_attention_wacv2020,Pal_semantic-gaze_cvpr2021,Baee_MEDIRL_iccv2021},
driver behavior change prediction~\cite{li2020make,li2023TPAMI}.
Although significant progress has been made, current evaluation strategies utilize independent datasets, which pose challenges in terms of making direct comparisons and impede collective efforts aimed at improving safety performance. 
For instance, collision 
anticipation
methods are evaluated on incidentDB~\cite{suzuki2018anticipating} or CTA datasets~\cite{You_eccv2020}, whereas trajectory prediction-based methods~\cite{Trajectron_iccv2019, Trajectron++_eccv2020,mantra,zhou2022hivt,nayakanti2023wayformer,zhou2023query} are evaluated on preprocessed trajectory datasets~\cite{nuscenes,waymo_motion_2021,zhan2019interaction}.
Additionally, object importance or risk region prediction-based methods use separate datasets~\cite{Spain_importance_eccv2008,Ohn-Bar_object_importance_2017,Gao_goal_icra2019,Zhang_graph_important_icra2020}.
%
There is currently an absence of a standardized benchmark for comprehensive evaluation of different risk identification algorithms.

%
%
%

We present RiskBench, a benchmark designed for risk identification, with a specific focus on assessing risks stemming from dynamic traffic participants (vehicles, pedestrians, and motorcyclists) and unexpected events (collisions and construction zones) through a scenario-based methodology.
This approach is prevalent in the automotive field~\cite{streetwise,Menzel_IV2018,nhtsa}, enabling comprehensive scenario-based performance analysis~\cite{scenariosurvey,SafeBench_neurips2020}.
We design a scenario taxonomy that enables the methodical collection of ground truth risks.
%
The taxonomy comprises four interaction types: \textit{Interactive}: yielding to dynamic risks~\cite{carlachallenge2022,SafeBench_neurips2020,agarwal2023ordered,kung2023action}, \textit{Collision}: crashing scenario~\cite{chan2016anticipating,fang2019dada,you2020cta}, \textit{Obstacle}: interacting with static elements~\cite{carlachallenge2022,nuscenes}, and \textit{Non-interactive}: normal driving~\cite{CordtsCVPR2016Cityscapes,bdd100k,Geiger2012CVPR,nuscenes}, aiming to cover diverse risks explored in the community. 
%
The four interaction types are depicted in Fig.~\ref{risk}.

\begin{table*}[t!]
    \scriptsize
    \centering
        \caption{\textbf{Comparison with traffic scene datasets and benchmarks relevant to risk identification and scenario-based analysis.} In scenario types, I, C, O, and N represent \textit{Interactive}, \textit{Collision}, \textit{Obstacle}, and \textit{Non-interactive}, respectively. For road topology, F, T, S, R, and \xmark \, represent \textit{4-way intersection}, \textit{T-intersection}, \textit{Straight}, \textit{Roundabout}, and \textit{Not Categorized}, respectively. For Risk Identification metrics, L, A, C, and P represent Risk Localization, Risk Anticipation, Temporal Consistency, and Planning Awareness, respectively. The column, Sensors, involves V and T, which stand for Vision and Trajectory, respectively.}
        \vspace{-2mm}
        \begin{tabular}{c|c|c|c|c|c}
            \multirow{2}*{ \begin{tabular}{@{}l@{}}  \textbf{Dataset} \end{tabular}} 
            & 
            \multirow{2}*{ \begin{tabular}{@{}c@{}} \textbf{\# of Categorized} \\ \textbf{Scenarios}  \end{tabular}} 
            &
            \multirow{2}*{ \begin{tabular}{@{}c@{}}           \textbf{Scenario Types} \\ 
            \begin{tabular}{{p{.075cm}p{.075cm}p{.075cm}p{.075cm}}} I& C & O &N \end{tabular}  \end{tabular}}
            & 
            \multirow{2}*{ \begin{tabular}{@{}@{}c@{}@{}}              \textbf{Road Topology} \\ 
            \begin{tabular}{{p{.075cm}p{.075cm}p{.075cm}p{.075cm}}} F &T & S & R \end{tabular} \end{tabular}} 
            &
            \multirow{2}*{ \begin{tabular}{@{}@{}c@{}@{}}              \textbf{Risk Id. Metrics} \\ \begin{tabular}{{p{.075cm}p{.075cm}p{.075cm}p{.075cm}}} L &A &C & P \end{tabular} \end{tabular}}
             &
            \multirow{2}*{ \begin{tabular}{@{}c@{}}              \textbf{Sensors} \\ \begin{tabular}{{p{.05cm}p{.05cm}}} V & T \end{tabular} \end{tabular}}
             \\
            & &  & & \\ 
            \hline
            \textbf{RiskBench}& 
            \begin{tabular}{@{}l@{}} 6916 \end{tabular} &
            \begin{tabular}{{p{.075cm}|p{.075cm}|p{.075cm}|p{.075cm}}}\checkmark&\checkmark&\checkmark&\checkmark \end{tabular} &
            \begin{tabular}{{p{.075cm}|p{.075cm}|p{.075cm}|p{.075cm}}}\checkmark&\checkmark&\checkmark&\checkmark \end{tabular}&
            \begin{tabular}{{p{.075cm}|p{.075cm}|p{.075cm}|p{.075cm}}}\checkmark&\checkmark&\checkmark&\checkmark \end{tabular}&
            \begin{tabular}{{p{.05cm}|p{.05cm}}} \checkmark&\checkmark \end{tabular} \\
            
            CARLA Challenge~\cite{carlachallenge2022} & 
            23   & 
            \begin{tabular}{{p{.075cm}|p{.075cm}|p{.075cm}|p{.075cm}}}\checkmark&&\checkmark&\checkmark \end{tabular}  & 
            \begin{tabular}{{p{.075cm}|p{.075cm}|p{.075cm}|p{.075cm}}} \checkmark&\checkmark&\checkmark& \end{tabular} &
            \begin{tabular}{{p{.075cm}|p{.075cm}|p{.075cm}|p{.075cm}}} &&&\checkmark \end{tabular} &
            \begin{tabular}{{p{.05cm}|p{.05cm}}} \checkmark&\checkmark \end{tabular} \\

            SafeBench~\cite{SafeBench_neurips2020} & 
            8 & 
            \begin{tabular}{{p{.075cm}|p{.075cm}|p{.075cm}|p{.075cm}}} \checkmark&&\checkmark&\checkmark \end{tabular} & 
            \begin{tabular}{{p{.075cm}|p{.075cm}|p{.075cm}|p{.075cm}}} \checkmark&\checkmark&\checkmark& \end{tabular} &
            \begin{tabular}{{p{.075cm}|p{.075cm}|p{.075cm}|p{.075cm}}} &&&\checkmark \end{tabular} &
            \begin{tabular}{{p{.05cm}|p{.05cm}}} &\checkmark \end{tabular}   \\

            Longest6~\cite{Chitta2022PAMI} & 
            7 & 
            \begin{tabular}{{p{.075cm}|p{.075cm}|p{.075cm}|p{.075cm}}} \checkmark&&& \checkmark \end{tabular} & 
            \begin{tabular}{{p{.075cm}|p{.075cm}|p{.075cm}|p{.075cm}}} \checkmark&&\checkmark& \end{tabular} &
            \begin{tabular}{{p{.075cm}|p{.075cm}|p{.075cm}|p{.075cm}}} &&&\checkmark \end{tabular} &
            \begin{tabular}{{p{.05cm}|p{.05cm}}} \checkmark&\checkmark \end{tabular}   \\

            DOS~\cite{shao2023reasonnet}& 100 &
            \begin{tabular}{{p{.075cm}|p{.075cm}|p{.075cm}|p{.075cm}}} \checkmark& & & \end{tabular}&
            \begin{tabular}{{p{.075cm}|p{.075cm}|p{.075cm}|p{.075cm}}} \checkmark&  & \checkmark& \end{tabular} &
            \begin{tabular}{{p{.075cm}|p{.075cm}|p{.075cm}|p{.075cm}}}& & &\checkmark  \end{tabular} &
            \begin{tabular}{{p{.05cm}|p{.05cm}}} \checkmark&  \end{tabular}\\

            GTACrash~\cite{kim2019aaai} & 
            N/A & 
            \begin{tabular}{{p{.075cm}|p{.075cm}|p{.075cm}|p{.075cm}}} \checkmark&\checkmark& &\checkmark \end{tabular} & 
            \begin{tabular}{{p{.075cm}|p{.075cm}|p{.075cm}|p{.075cm}}} \xmark & \xmark & \xmark & \xmark \end{tabular} &
            \begin{tabular}{{p{.075cm}|p{.075cm}|p{.075cm}|p{.075cm}}} \checkmark& && \end{tabular} &
            \begin{tabular}{{p{.05cm}|p{.05cm}}} \checkmark& \end{tabular}\\

            nuPlans~\cite{Caesar2021nuPlanAC} & 
            75 & 
            \begin{tabular}{{p{.075cm}|p{.075cm}|p{.075cm}|p{.075cm}}} \checkmark & & \checkmark &\checkmark \end{tabular}  & 
            \begin{tabular}{{p{.075cm}|p{.075cm}|p{.075cm}|p{.075cm}}} \checkmark &\checkmark & \checkmark &\checkmark \end{tabular} &
            \begin{tabular}{{p{.075cm}|p{.075cm}|p{.075cm}|p{.075cm}}}& &&\checkmark \end{tabular} &
            \begin{tabular}{{p{.05cm}|p{.05cm}}} &\checkmark \end{tabular}\\ 
            
            DAD~\cite{chan2016anticipating} & 
            4 & 
            \begin{tabular}{{p{.075cm}|p{.075cm}|p{.075cm}|p{.075cm}}} \checkmark&\checkmark&&\checkmark\end{tabular} & 
            \begin{tabular}{{p{.075cm}|p{.075cm}|p{.075cm}|p{.075cm}}} \xmark & \xmark & \xmark & \xmark \end{tabular} &
            \begin{tabular}{{p{.075cm}|p{.075cm}|p{.075cm}|p{.075cm}}}\checkmark&\checkmark&& \end{tabular} &
            \begin{tabular}{{p{.05cm}|p{.05cm}}} \checkmark& \end{tabular}\\ 
            
            DADA~\cite{fang2019dada} &  
            2,000 & 
            \begin{tabular}{{p{.075cm}|p{.075cm}|p{.075cm}|p{.075cm}}} &\checkmark&& \end{tabular} &  
            \begin{tabular}{{p{.075cm}|p{.075cm}|p{.075cm}|p{.075cm}}} \xmark & \xmark & \xmark & \xmark \end{tabular} &
            \begin{tabular}{{p{.075cm}|p{.075cm}|p{.075cm}|p{.075cm}}}\checkmark&\checkmark& &\end{tabular} &
            \begin{tabular}{{p{.05cm}|p{.05cm}}} \checkmark& \end{tabular}\\
            
            CTA~\cite{you2020cta} & 
            18 & 
            \begin{tabular}{{p{.075cm}|p{.075cm}|p{.075cm}|p{.075cm}}} \checkmark&\checkmark&\checkmark& \end{tabular} & 
            \begin{tabular}{{p{.075cm}|p{.075cm}|p{.075cm}|p{.075cm}}} \xmark & \xmark & \xmark & \xmark \end{tabular} &
            \begin{tabular}{{p{.075cm}|p{.075cm}|p{.075cm}|p{.075cm}}}\checkmark&\checkmark&& \end{tabular}&
            \begin{tabular}{{p{.075cm}|p{.075cm}}} \checkmark& \end{tabular}\\

            BDD-A~\cite{Xia_ACCV_2018} & 
            N/A & 
            \begin{tabular}{{p{.075cm}|p{.075cm}|p{.075cm}|p{.075cm}}} \checkmark& & &\checkmark \end{tabular} & 
            \begin{tabular}{{p{.075cm}|p{.075cm}|p{.075cm}|p{.075cm}}} \xmark & \xmark & \xmark & \xmark \end{tabular} &
            \begin{tabular}{{p{.075cm}|p{.075cm}|p{.075cm}|p{.075cm}}}\checkmark& & &\  \end{tabular} &
            \begin{tabular}{{p{.05cm}|p{.05cm}}} \checkmark&  \end{tabular}\\

            DRAMA~\cite{malla2023drama} & 
            17785 &
            \begin{tabular}{{p{.075cm}|p{.075cm}|p{.075cm}|p{.075cm}}} \checkmark& & \checkmark & \end{tabular}&
            \begin{tabular}{{p{.075cm}|p{.075cm}|p{.075cm}|p{.075cm}}} \checkmark &  & \checkmark & \end{tabular} &
            \begin{tabular}{{p{.075cm}|p{.075cm}|p{.075cm}|p{.075cm}}}\checkmark& & &  \end{tabular} &
            \begin{tabular}{{p{.05cm}|p{.05cm}}} \checkmark&  \end{tabular}\\

            TrafficNet~\cite{ding2017trafficnet} & 
            30  &
            \begin{tabular}{{p{.075cm}|p{.075cm}|p{.075cm}|p{.075cm}}}  \checkmark&&&\checkmark \end{tabular}  &
            \begin{tabular}{{p{.075cm}|p{.075cm}|p{.075cm}|p{.075cm}}} \xmark & \xmark & \xmark & \xmark \end{tabular} &
            \begin{tabular}{{p{.075cm}|p{.075cm}|p{.075cm}|p{.075cm}}} &&&\checkmark \end{tabular} &
            \begin{tabular}{{p{.05cm}|p{.05cm}}} & \end{tabular} \\
            
            \bottomrule

        \end{tabular}

        \label{table:dataset_comparison}
        \vspace{-.2cm}
\end{table*}

We construct the RiskBench dataset in the CARLA simulator~\cite{Dosovitskiy17carla}, resulting in \textbf{6916} scenarios from 4 interaction types, 6 actor-behavior categories, 14 different maps, 237 unique road structures, 4 traffic violations, 3 traffic densities, and 21 weather and lighting conditions.
%
%
Simulation-based benchmarking has been widely adopted in both the research community~\cite{Dosovitskiy17carla,kim2019aaai,SafeBench_neurips2020,Prakash2021CVPR,li2022v2x,shift2022,xu2022v2xvit,li2023towards,shao2023reasonnet} and industry (e.g., Waymo~\cite{waymo2021} and Waabi~\cite{yang2023unisim}).
For instance, SafeBench~\cite{SafeBench_neurips2020} generates safety-critical scenarios in the CARLA simulator for evaluating the performance of driving models. 
Similarly, DOS~\cite{shao2023reasonnet} also assesses driving models within the CARLA environment, placing particular emphasis on the analysis of various occlusion events.
%

%
%

%
Riskbench mitigates the following challenges when it comes to benchmarking risk identification algorithms in real-world settings: 
First, simulation serves as a crucial preliminary phase preceding costly real-world testing and deployment. It aids researchers and developers in pinpointing critical flaws and vulnerabilities.
Second, current large-scale real-world datasets are collected naturalistically~\cite{nuscenes,bdd100k,vasili2018hdd,waymo_motion_2021}. They lack the capacity to regulate the occurrence frequency of various scenarios~\cite{kung2023action}. 
%
This limitation hampers the thorough scenario-based performance analysis.
%
%
Third, static real-world datasets are inadequate for supporting interactive evaluation, which is essential for validating the impact of risk identification on subsequent planning processes.

We devise three metrics to assess the performance of a risk identification algorithm, drawing inspiration from the ISO26262 Hazard and Risk Assessment Methodology~\cite{iso26262}.
Specifically, we evaluate algorithm's ability to (1) localize risks, (2) anticipate risks, and (3) facilitate decision-making. 
The first two metrics are well-established in collision detection and anticipation literature~\cite{chan2016anticipating,suzuki2018anticipating,you2020cta,badki2021binary}.
Our contribution lies in the third metric, which is a planning-aware metric called Influenced Ratio (IR). 
The key insight is that if the identified risks are accurate, the planner should possess the ability to generate a collision-free path. 
On the contrary, if a model makes an inaccurate identification, the planner could potentially guide the ego vehicle into a situation with a higher risk of collision or near-collision. 

We benchmark ten risk identification algorithms, including rule-based, trajectory prediction and collision checking-based, collision 
anticipation,
and behavior prediction-based algorithms.
%
Our scenario-based performance analysis shows that existing risk identification algorithms demand considerable endeavor. We identify two crucial perspectives for future research: (1) ensuring temporal consistency of risk identification and (2) improving object-centric representation learning (Please see Video Submission for details).

Our contributions are summarized as follows:
\begin{itemize}
  \item We introduce \textbf{RiskBench}, 
  a scenario-based benchmark of identifying risks induced by traffic participants and unexpected events, alongside a scenario taxonomy for methodical collection of ground truth risks.
  \item We develop a scenario-based benchmark to systematically evaluate the ability of risk identification algorithms to (1) localize risks, (2) anticipate risks, and (3) facilitate decision-making. 
  \item We conduct extensive scenario-based analysis on ten algorithms and discuss the limitations of existing risk identification algorithms.  
\end{itemize}

\section{Related Work}
\label{related}

\noindent\textbf{Traffic Scene Datasets and Benchmarks.}
Various datasets and benchmarks have been released in recent years for object detection and tracking~\cite{Geiger2012CVPR,360LiDARTracking_ICRA_2019,nuscenes},
%
%
and activity recognition and intention prediction of traffic participants~\cite{rasouli2017they,vasili2018hdd,rasouli2019pie,titan_CVPR2020,girase2021loki}. 
%
We observe substantial progress because of the standardized benchmarks.
There are independent efforts tailored for risk identification, such as collision anticipation~\cite{chan2016anticipating,suzuki2018anticipating,You_eccv2020,fang2019dada}, risk region prediction~\cite{Herzig_iccvw2019}, object importance~\cite{Spain_importance_eccv2008,Ohn-Bar_object_importance_2017,Gao_goal_icra2019},
trajectory prediction and collision checking~\cite{waymo_motion_2021,Caesar2021nuPlanAC},
human gaze prediction~\cite{Alletto_Dreye_cvprw2016,Xia_ACCV_2018,Pal_semantic-gaze_cvpr2021,Baee_MEDIRL_iccv2021},
risk localization and captioning~\cite{malla2023drama},
behavior change detection and causal inference~\cite{vasili2018hdd}.
%
Additionally, scenario-based safety assessment benchmarks are explored~\cite{SafeBench_neurips2020,carlachallenge2022}.
Table~\ref{table:dataset_comparison} provides a summary of datasets and benchmarks that are relevant to risk identification and scenario-based performance analysis.
%

%
Our work complements existing datasets and benchmarks by constructing a common benchmark that entails different definitions of risks explored in the community.
%
%
%
In addition, we propose a comprehensive evaluation protocol, including the planning-aware metric, which has been explored. 
SafeBench~\cite{SafeBench_neurips2020} is the most relevant work among existing sets.
The authors apply safety-critical scenario generation algorithms to generate challenging testing cases to evaluate the safety and robustness of autonomous driving algorithms. 
%
In pursuit of collaborative efforts, we believe there is a great synergy to combine the initiatives described in~\cite{carlachallenge2022,SafeBench_neurips2020}.
%


\noindent\textbf{Planning-aware Metrics.}
The drawback of prevalent metrics (e.g., precision/recall) used in perception algorithms is that they penalize all incorrect detection equally, without considering the downstream tasks.
%
Recently, planning-aware metrics have been explored for perception and trajectory forecasting in the community~\cite{planning-aware-cvpr2020,plant_corl2022}.
Philion et al.,~\cite{planning-aware-cvpr2020} design the planning-aware metric for perception through training a robust planner, using the nuScenes dataset~\cite{nuscenes},
to plan a driving trajectory based on its semantic observations (detection).
%
%
%
Recently, Renz et al.,~\cite{plant_corl2022} proposed the Relative Filtered Driving Score to quantify the ability of planners to identify relevant objects. The concept is highly relevant to the proposed planning-aware metric, IR. 
%
We design IR to evaluate the ability of risk identification algorithms instead of planners. 
It will be our future work to consider the two tasks jointly as they go hand in hand.
%


 

\section{The \textbf{RiskBench} Dataset}
\label{sec: scenario_generation}

\begin{figure}[t!]
\centering
\includegraphics[width=0.9\columnwidth, trim={0 0cm 0 0cm},clip]{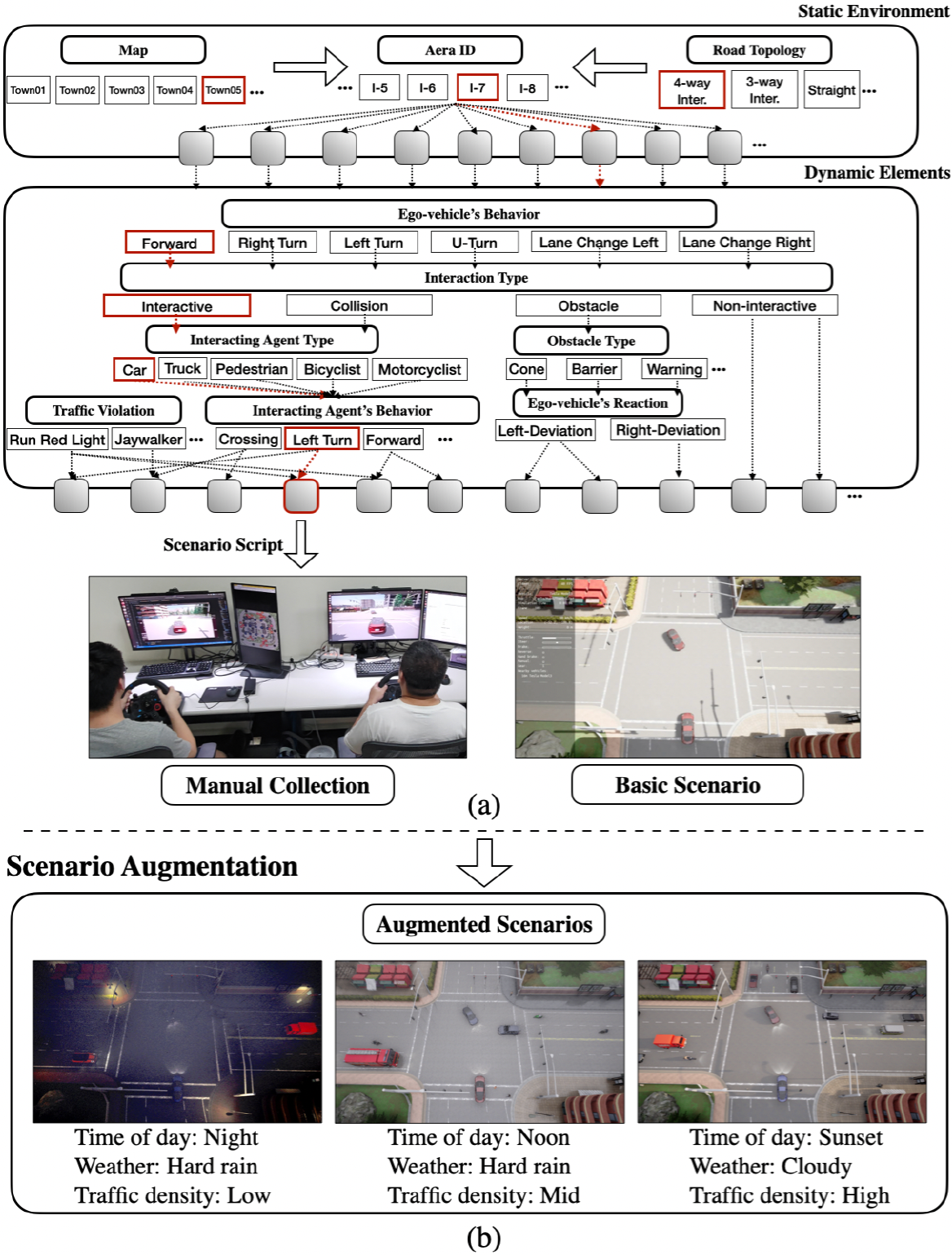} 
\caption{\textbf{Scenario Collection Pipeline.}
%
The scenario taxonomy is designed to enable a systematic collection of ground truth risks induced by dynamic traffic participants and unexpected events.
The taxonomy includes various attributes such as road topology, scenario types, ego vehicle behavior, and traffic participants' behavior. From this taxonomy, if a scenario script is set, two human subjects can act accordingly.
%
To form the final scenario dataset, we augment the collected scenario by changing attributes, including time of day, weather conditions, and traffic density.}
\label{pipeline}
\vspace{-0.5cm}
\end{figure}

\begin{figure*}[t!]
\centering
\includegraphics[width=1.0\textwidth]{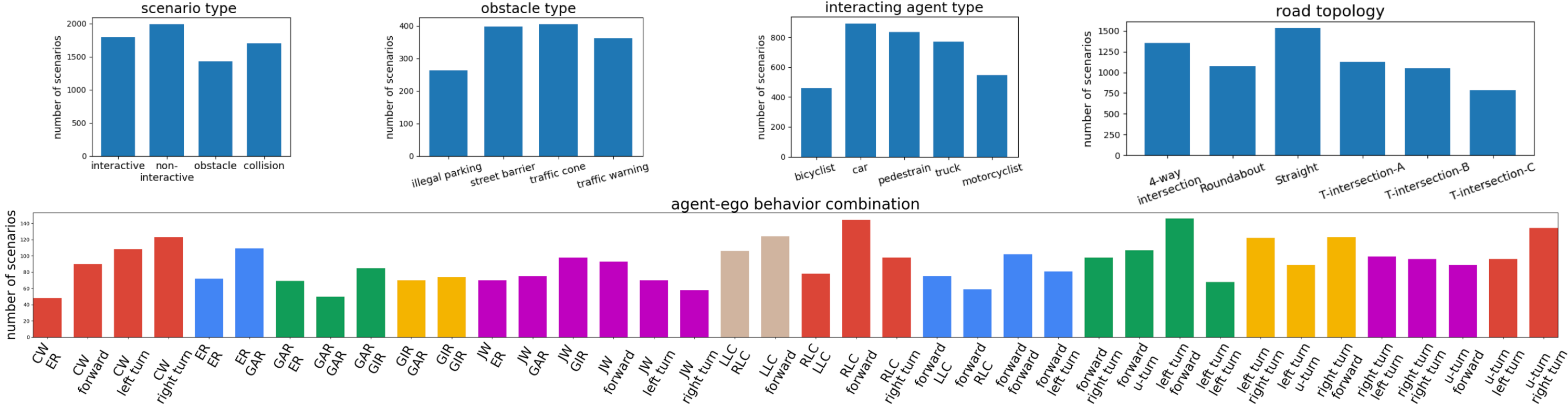}

\caption{\textbf{The RiskBench Dataset Statistics.} We denote \textit{Crosswalking} as CW, \textit{Jay-walking} as JW, \textit{Left Lane Change} as LLC, \textit{Right Lane Change} as RLC, \textit{Go into Roundabout} as GIR, \textit{Exit Roundabout} as ER, and \textit{Go Around Roundabout} as GAR.}
\label{dataset}
\vspace{-10pt}
\end{figure*}

This section presents the scenario taxonomy, and the data collection and augmentation pipeline.

\subsection{The Taxonomy of Scenarios}
The motivation for designing a scenario taxonomy is to enable a systematic collection of \textbf{ground truth risks} induced by dynamic traffic participants and unexpected events.

\noindent\textbf{Static Environment} contains the following attributes, i.e., \textit{Map}, \textit{Area ID}, and \textit{Road Topology}. 
The attribute \textit{Map} has \textit{8} towns constructed in the CARLA ecosystem. 
We assign all road segments and junctions with a \textit{Area ID}. In total, we have unique \textit{231} IDs.
\textit{Road topology} denotes the road structure of each \textit{Area ID}, e.g., 4-way intersection. 
Note that, we reconstruct real-world maps from CAROM-Air~\cite{LuEWWCWYY23} in the CARLA simulator because the simulator provides a limited number of roundabouts. 
Note that, we select 6 out of 13 semantic maps provided by CAROM-Air.
%

\noindent\textbf{Dynamic Elements} 
involve the following attributes, i.e., \textit{Interaction Type}, \textit{Interacting Agent Type}, \textit{Interacting Agent's Behavior}, and \textit{Ego-Vehicle's Behavior}.
To collect a scenario, we first set a value for \textit{Ego-vehicle's Behavior} then determine the value of \textit{Interaction Type}.
If the value of \textit{Interaction Type} is \textit{Interactive} and \textit{Collision}, we first set a value for \textit{Interacting Agent Type} and \textit{Interacting Agent's Behavior}.
Note that the assignment to \textit{Interacting Agent Type} and \textit{Interacting Agent's Behavior} define the type and behavior of the \textbf{ground truth risk} in a scenario. 
Then we set a value for \textit{Interacting Agent Behavior}.
%
If the value of \textit{Interaction Type} is \textit{Obstacle}, we first set a value for \textit{Obstacle Type}. 
Note that the assignment to \textit{Obstacle Type} defines the type of the \textbf{ground truth risk}.
Then we set a value for \textit{Ego-vehicle's reaction}. 
Please refer to our website for the full description of the values of each attribute. 
%
%
%
%

\subsection{Data Collection} 
%
%
Fig.~\ref{pipeline} depicts the overall pipeline.
%
We script a scenario by specifying attribute values.
%
For instance, we specify the values of \textit{Map}, \textit{Area ID}, \textit{Road Topology}, \textit{Ego-vehile's Behavior}, \textit{Interaction Type}, \textit{Interacting Agent Type}, and \textit{Interacting Agent's Behavior} as \textit{Town05}, \textit{I-7}, \textit{4-way intersection}, \textit{Forward}, \textit{Interactive}, \textit{Car}, and \textit{Left Turn}, respectively.
To authenticate interactions and address the rarity of scenarios, we enlist two human subjects for script execution.
%
One participant assumes the role of the ego vehicle, while the other takes on the role of the interacting agent, i.e., acting as the ground truth risk.
%

A Basic Scenario collection begins with both subjects' movement and concludes upon interaction completion or the ego vehicle's completion of its assigned behavior.
We collect the trajectories of both agents and the actors' blueprints, defined by the CARLA simulator~\cite{Dosovitskiy17carla}.
%



\subsection{Scenario Augmentation}
To boost interaction complexity and visual challenges, we create a scenario augmentation pipeline.
%
Specifically, we replay a \textbf{Basic Scenario} in CARLA using the recorded trajectories of the ego vehicle and interacting agent. We apply three augmentation strategies, i.e., random actor injection, time-of-day, and weather condition adjustment, as illustrated in Fig. \ref{pipeline} (b). 
For random actor injection, we spawn random actors around the ego vehicle as the distraction and ensure their trajectories will not interact with the ego vehicle. 
We set the level of traffic density to be \textit{low}, \textit{medium}, or \textit{high}. 
%
%
For time-of-day and weather condition adjustment, we uniformly sample a value for time-of-day (i.e., noon, sunset, or night). We randomly select the cloudiness and rain intensity.
Our data collection pipeline enables the creation of a well-organized dataset and the statistics are reported in Fig.~\ref{dataset}.

\section{Risk Identification Benchmark}
We present risk identification algorithms and 
implementation details, metrics, and benchmarking results.


\subsection{Risk Identification Baselines} 
The baselines take a sequence of historical data as input and output a risk score for each road user (e.g., vehicle or pedestrian) or an unexpected event (e.g., collision or construction zone). 
We consider a road user or an unexpected event risk if the score exceeds a predefined threshold.
We implement 10 risk identification algorithms and categorize them into the following four types.\\
\noindent\textbf{1. Rule-based.} \textbf{Random:} We randomly pick an object. 
\textbf{Range:} We pick any objects within a predefined distance. \\ 
\noindent\textbf{2. Trajectory Prediction and Collision Checking.} We follow the protocol of~\cite{risk_assessment_Lefevre_ROBOMECH} to identify risk. Specifically, if there is an overlap between the ego vehicle‘s and a road user's future trajectories, the road user is the predicted risk. We select \textbf{Kalman filter}~\cite{thrun2002probabilistic}, adversarial network \textbf{Social-GAN}~\cite{social-gan}, memory augmented network \textbf{MANTRA}~\cite{mantra}, and recent query-based \textbf{QCNet}~\cite{zhou2023query}, which is 2$^{\mathtt{nd}}$ place winner of the Argoverse challenges~\cite{chang2019argoverse,argoverse2}.\\
%
\noindent\textbf{3. Collision Anticipation.} \textbf{DSA}~\cite{chan2016anticipating} and \textbf{RRL}~\cite{zeng2017agent} identify risks via collision prediction with exponential loss~\cite{jain2016exploss}, encouraging early anticipation of risk. 
Note that \textbf{RRL} uses ground truth risk supervision in its training.\\
\noindent\textbf{4. Behavior Prediction-based.} The key insight of Behavior-based is that a risky situation is identified if the ego vehicle behavior is influenced~\cite{li2020make}. We manually label the time index, where the ego vehicle's behavior is influenced in a scenario. 
We implement two baselines.  %
\textbf{Behavior Prediction (BP)}~\cite{li2020gcn} utilizes graph attention networks~\cite{gan} to model interactions between traffic participants and the ego-vehicle to predict ego behavior. If the predicted behavior is "Influenced," we select the object with the highest attention scores as the risk.
\textbf{Behavior Change Prediction (BCP)}~\cite{li2020make} identifies risk by applying a causal inference-based approach~\cite{li2020make}.
%
    %

It is worth noting that the collision anticipation and behavior-based approaches are vision-based.

\begin{table*}[t!]
    \small
    
    \centering
    \caption{\textbf{Risk Localization and Anticipation Evaluation.} The notations P, R, PIC, and FA denote precision, recall, progressive increasing cost, and false alarm rate, respectively.}
        \begin{tabular}
            { @{}l@{\;}@{\;} c @{\;}@{\;} @{\;}@{\;}c @{\;}@{\;} @{\;} c @{\;} @{\;} c @{\;} @{\;}@{\;} c @{\;} @{\;}@{\;} c @{\;}@{\;}@{\;} c @{\;}@{\;}@{\;} c @{\;} c @{\;} c @{\;} c @{\;}l@{\;}}
            \toprule
            \multirow{2}{*}{ \begin{tabular}{@{\;}c@{\;}} \end{tabular}} & 
             \multicolumn{3}{c}{Interactive}  & 
             \multicolumn{3}{c}{Collision} &
             \multicolumn{3}{c}{Obstacle} &  
             \multicolumn{1}{c}{Non-inter.} &
             \multicolumn{1}{c}{All} 
             \\
             \cmidrule(lr){2-4} \cmidrule(lr){5-7} \cmidrule(lr){8-10} \cmidrule(lr){11-11} \cmidrule(lr){12-12} 
             &
             \begin{tabular}{c@{\;}@{\;}}  P (\%)\end{tabular} & 
             \begin{tabular}{c@{\;}@{\;}@{\;}}  R (\%)\end{tabular} & 
             \begin{tabular}{c@{\;}@{\;}}  PIC\end{tabular} &
             \begin{tabular}{@{}c@{}@{}}  P (\%)  \end{tabular} & 
             \begin{tabular}{c@{\;}@{\;}}  R (\%)\end{tabular} & 
             \begin{tabular}{@{}c@{}@{}}  PIC   \end{tabular} & 
             \begin{tabular}{@{}c@{}@{}}  P (\%) \end{tabular} &
             \begin{tabular}{c@{\;}@{\;}}  R (\%)\end{tabular} & 
             \begin{tabular}{@{}c@{}@{}}  PIC  \end{tabular} &
             \begin{tabular}{@{}c@{}}  FA(\%)\end{tabular} &
             \begin{tabular}{@{}c@{}}  F-1
             \end{tabular}
            \\
             \midrule
            Random
             & 9.5   & 50.0 & 16.7
             & 19.2 & 49.8 & 15.7
             & 7.2   & 50.3 & 11.4
             & 80.8 & 15.1 \\
             Range (5m)
             & 38.7 & 10.7 & 20.3
             & 85.6 & 27.7 & 4.1
             & 45.0 & 18.3 & 10.2
             & 3.3 & 30.0 \\
             Range (10m)
             & 43.6 & \textbf{69.1} & \textbf{7.1}
             & 69.3 & 62.4 & 3.8
             & 33.3 & \textbf{88.5} & \textbf{0.8}
             & 15.2 & \textbf{53.6} \\
             \midrule
             Kalman filter~\cite{thrun2002probabilistic}
             & 37.0 & 54.0 & 15.4
             & 59.9 & 58.1 & 4.2
             & 35.3 & 61.2 & 10.2
             & 18.8 & 46.7\\
             
             Social-GAN~\cite{social-gan}   
             &  39.4& 57.7& 14.1
             &  60.8& 45.4 & 7.8
             &  36.9& 69.3& 4.0
             &  16.3 &  46.0 \\
             
             MANTRA~\cite{mantra}
             & 36.3 & 62.1  & 13.3
             & 58.3 & 45.4 & 7.1
             & 35.7 & 71.8 & 2.5
             & 16.5 & 45.4 \\

             QCNet~\cite{zhou2023query}
             & 40.0 & 56.6  & 14.1
             & 60.3 & 45.4 & 7.3
             & 38.2 & 68.9 & 4.4
             & 14.5 & 46.4 \\
             
            \midrule
             DSA~\cite{chan2016anticipating} 
             & \textbf{54.7} & 19.7 & 20.9
             & 79.4 & 45.9 & \textbf{3.7}
             & \textbf{54.2} & 47.2 & 17.5
             & \textbf{3.3} & 46.8 \\
             RRL~\cite{zeng2017agent}        
             & 49.4 & 15.4 & 22.4 
             & \textbf{87.3} & \textbf{64.6} & 4.6 
             & 47.1 & 19.0 & 16.0 
             & 5.1 & 48.6 \\
             \midrule
             BP~\cite{li2020gcn}    
             & 36.4 & 16.3 & 25.6
             & 74.4 & 9.8 & 25.1
             & 22.4 & 3.0 & 27.9
             & 4.7   & 16.6 \\
             BCP~\cite{li2020make}
             & 48.7 & 29.2 & 22.8
             & 76.8 & 20.7 & 20.8
             & 32.1 & 35.2 & 20.5
             & 4.7   & 33.4 \\
             \midrule
            \bottomrule
        \end{tabular}
        \label{table:risk_assessmen_benchmark}
        \vspace{-.2cm}
\end{table*}

\subsection{Evaluation Metrics}
We devise three metrics that evaluate the ability of a risk identification algorithm to (1) identify locations of risks, (2) anticipate risks, and (3) facilitate decision-making.
%
%

\noindent\textbf{1. Risk Localization}: We utilize \textbf{precision} and \textbf{recall} for risk localization evaluation, following the existing collision and anticipation literature~\cite{chan2016anticipating,suzuki2018anticipating,You_eccv2020}. 
%
%
%
%
We utilize the \textbf{False Alarm Rate} for \textit{Non-interactive} scenarios (no risky dynamic elements or events), i.e., a positive prediction is a false alarm.
%
%

\begin{figure}[t!]
\centering
\includegraphics[width=0.75\columnwidth]{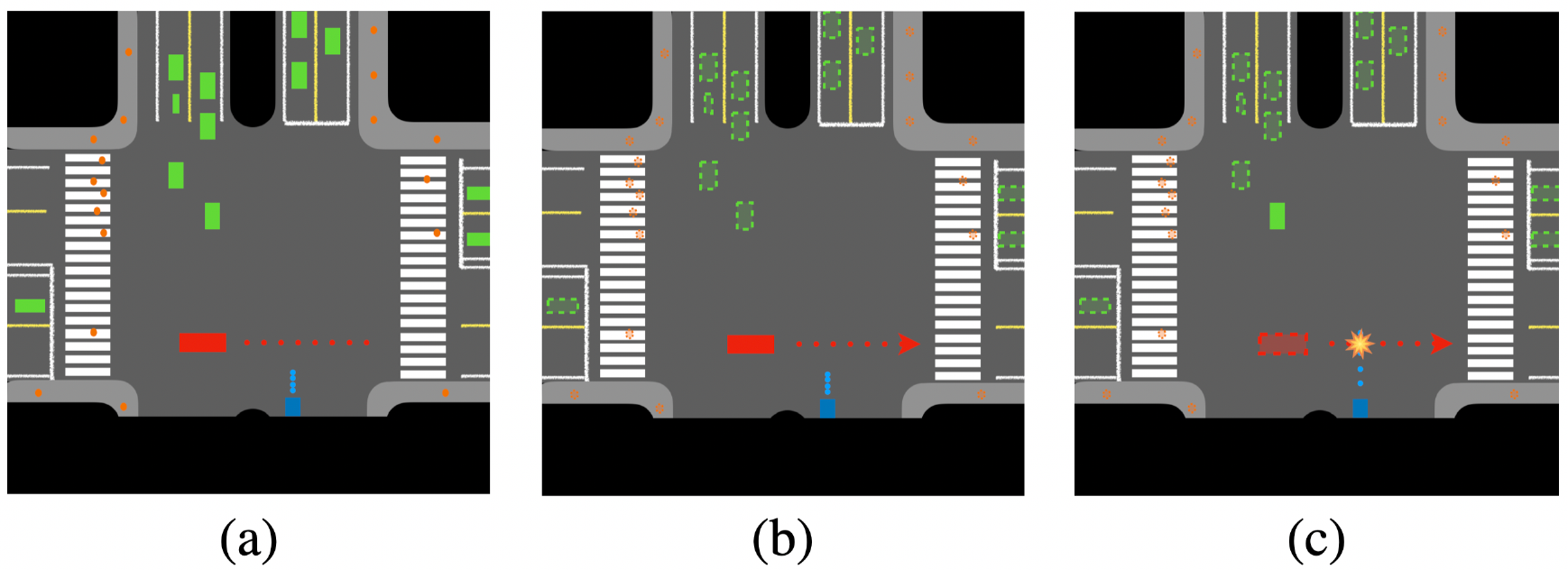}
\caption{\textbf{Planning-aware Metric.} 
An ideal planner should be able to yield to the crossing vehicle as shown in (a). If a risk identification model successfully identifies the vehicle as the risk, we hide all the other objects. The planner should plan a slow-down path, as shown in (b). In contrast, if the model identifies the wrong object as risk (e.g., the green one), the planner will plan a path to (nearly) collide with the 
\textbf{True Risk}, as shown in (c).}

\label{planning}
\vspace{-.5cm}
\end{figure}

\noindent\textbf{2. Risk Anticipation:}
%
Risk Localization treats the prediction of each frame independently, without evaluating if a model can anticipate risks or unexpected events.
Time-to-collision (TTC)~\cite{chan2016anticipating,suzuki2018anticipating,badki2021binary} is one of the most common metrics to evaluate risk anticipation.
However, they do not consider false identification of risk. 
To this end, we propose \textbf{Progressive Increasing Cost} (PIC), an anticipation and localization-aware metric. 
%
%
Our insight is that we penalize a false identification more when we are closer to the risk than far from the risk.
We propose to combine the notion of TTC, the exponential loss~\cite{jain2016exploss}, and the F1-score to establish an anticipation and localization-aware metric.
Given a scenario with ${T}$ frames, we calculate the F-1 score for each frame ${t}$ and define the metric as $\textrm{PIC} = -\sum_{t=1}^{n} e^{-(T-t)}\mathtt{log}(F1_{t})$.


%
\noindent\textbf{3. Planning Awareness:} We propose \textbf{Influenced Ratio} (IR) to evaluate if a risk identification algorithm can facilitate decision-making. 
%
%
%
Assume we access an ideal planner that can interact with traffic participants and unexpected events. Our insight is that, if the identified risks are accurate, the planner should plan a path without 
collisions, as illustrated in Figure~\ref{planning} (b).
%
%
%
%
Specifically, IR is defined as $\textrm{IR}  = |D_{orig} - D_{post}| / D_{orig}$,
where the parameter $D_{orig}$ denotes the closest distance while interacting between the ego vehicle and the true risk in a scenario using the full observation. The parameter $D_{post}$ denotes the closest distance while interacting between the ego vehicle and the true risk with the masked observation. The value of \textrm{IR} is between 0 and 1. If the value is closer to 1, there is a higher possibility that there is a false identification influencing the planner's decision-making, resulting in safety-critical situations. We also adopt \textbf{Collision Rate}, a commonly used metric for evaluating planning and decision-making~\cite{Caesar2021nuPlanAC,carlachallenge2022}. %

\subsection{Implementation Details}

\noindent \textbf{Dataset.}
The testing set comprises a total of 1689 scenarios, including 521 interactive, 375 collision, 322 obstacle, and 471 non-interactive scenarios.
We split train/val/test according to maps provided by CARLA~\cite{Dosovitskiy17carla}. 
%
%

\noindent \textbf{Inputs.} 
To make a fair comparison among all methods, vision-based algorithms (i.e., \textbf{DSA}~\cite{chan2016anticipating}, \textbf{RRL}~\cite{zeng2017agent}, \textbf{BP}~\cite{li2020gcn}, and \textbf{BCP}~\cite{li2020make}) access the ground-truth tracklets as input. We restrict the field of view of all methods to the front view to eases the multi-view fusion problem.




\noindent \textbf{Planning-aware Evaluation.}
Our planner is the privileged model proposed in~\cite{chen2020lbc}, which takes a bird's-eye-view of semantic maps as input and outputs waypoints. 
%
The model is trained to imitate demonstrations collected in the RiskBench. 
The evaluation is conducted on \textit{Interactive} and \textit{Obstacle} types. We exclude \textit{Collision} scenarios because we found the trained planner cannot handle collision scenarios and leave it for future work. 

\subsection{Risk Localization and Anticipation Evaluation}
\label{subsec: benchmarking}

We evaluate algorithms' ability to localize risks via Precision, Recall, and False Alarm Rate (FA rate) and to anticipate risk via Progressive Increasing Cost (PIC). 
%
%
We also report an overall F-1 score, considering all scenarios. 
The complete scenario-based performance analysis is shown in Table \ref{table:risk_assessmen_benchmark}. 

\noindent\textbf{Rule-based.}
We find that \textbf{Range (5m)} identifies risks only when the risk is very close, resulting in unsatisfactory recall in all safety-critical scenarios. 
%
%
Although increasing the distance threshold to 10m can significantly improve performance in most scenarios, \textbf{Range (10m)} increases FA in \textit{Non-interactive} scenarios compared to \textbf{5m}. Hyperparameter tuning for rule-based methods to strike a balance between precision and recall across diverse scenarios is challenging.\\ 
%
%
%
In collision scenarios, \textbf{Trajectory Prediction and Collision Checking} outperforms the other interaction types in terms of precision and PIC.
%
This is due to the short-term prediction horizon of these methods when the ego-vehicle decelerates, preventing trajectory overlap. 
This is because the predicted trajectories become short when the ego-vehicle slows down, which normally results in no overlap of trajectories. 
The same trend is seen in "Non-interactive" scenarios, where all trajectories remain fast, resulting in inferior FA.
\\
\noindent \textbf{Collision Anticipation.}
RRL outperforms all learning-based 
models in terms of the 
F-1 score on all scenarios 
since the method directly learns to localize 
risks.
We notice \textbf{RRL} and \textbf{DSA} do not perform well in \textit{Interactive} and \textit{Obstacle} scenarios because they are intrinsically different from \textit{Collision}.\\ 
\textbf{Behavior Prediction-based.} 
\textbf{BP}~\cite{li2020gcn} faces difficulty identifying the risk affecting ego-vehicle's action.
\textbf{BCP}~\cite{li2020make} performs better due to the causal inference.
However, the two methods generally have low recall due to weak supervision.   
%


\begin{table}[t!]
    \scriptsize
    \centering
        \caption{\textbf{Planning-aware Evaluation.} IR and CR denote \textbf{Influenced Ratio} and \textbf{Collision Rate}, respectively. 
        }
        \begin{tabular}
            {@{}l @{\;} @{\;} c @{\;}@{\;} @{\;}@{\;}c @{\;}@{\;} @{\;} c @{\;} @{\;} c @{\;} c @{\;} }
            \toprule
            \multirow{2}{*}{ \begin{tabular}{@{\;}c@{\;}} Risk ident. Algorithm \end{tabular}} & 
             \multicolumn{2}{c}{Interactive}  & 
             \multicolumn{2}{c}{Obstacle}  & 

             \\
             \cmidrule(lr){2-3} \cmidrule(lr){4-5}
             &
             \begin{tabular}{c@{\;}}  IR  \end{tabular} & 
             \begin{tabular}{c@{\;}}  CR (\%)  \end{tabular} & 
             \begin{tabular}{c@{\;}}  IR   \end{tabular} & 
             \begin{tabular}{c@{\;}}  CR (\%)  \end{tabular} & 
             \\
             \midrule
             Auto-pilot~\cite{Dosovitskiy17carla}
             & 0.45  & 33.6  & 0.52 & 58.7 \\
             Full observation
             & 0.00 &  0.0&  0.00& 0.0  \\
             Ground truth risk
             & 0.02 &  0.4&  0.04 & 5.3 \\
             \midrule
             Random
             & 0.11&  7.7 &  0.44 & 48.1 \\
             Range (10m)
             & 0.01 & 6.2   &  \textbf{0.24}& \textbf{26.4} \\
             \midrule
             Kalman Filter~\cite{thrun2002probabilistic}
             & 0.06 & 6.9 &  0.33 & 41.8 \\
             Social-GAN~\cite{social-gan}   
             & 0.16   & 10.0   & 0.49 & 55.3 \\
             MANTRA~\cite{mantra}   
             &  0.15  & 9.3   & 0.51 & 56.7 \\
             QCNet~\cite{zhou2023query}   
             &  0.05  &  10.0   & 0.52  &  57.2  \\
             \midrule
             DSA~\cite{chan2016anticipating}
             &  \textbf{0.01}  & \textbf{0.8}   & 0.37 & 38.5 \\
             RRL~\cite{zeng2017agent}
             &  0.01  & 1.2  & 0.49 &  54.3 \\
             \midrule
             BP~\cite{li2020gcn}
             & 0.14   & 9.3  & 0.41 & 47.1 \\
             BCP~\cite{li2020make}
             & 0.14  & 9.7  & 0.34 &  39.4 \\
             \midrule
            \bottomrule
        \end{tabular}

        \label{table:benchmarkplan}
\end{table}

\begin{table}[t!]
    \scriptsize
    \centering
\caption{\textbf{Temporal Consistency} is calculated within 1, 2, and 3 seconds before critical/collision points. 
}
\begin{tabular}
            {@{}l@{\;} @{\;} @{\;} c @{\;}@{\;}@{\;} @{\;}@{\;}@{\;}c @{\;}@{\;}@{\;} @{\;}@{\;}@{\;} c @{\;}@{\;}@{\;} @{\;}@{\;}@{\;} c @{\;} c @{\;}@{\;}@{\;} }
            \toprule
            \multirow{1}{*}{ \begin{tabular}{@{\;}c@{\;}} \end{tabular}} & 
             \multicolumn{1}{c}{1s}  & 
             \multicolumn{1}{c}{2s}  & 
             \multicolumn{1}{c}{3s}   & 
             &
             \\
             \midrule
             
             QCNet~\cite{zhou2023query}   
             & \textbf{50.2\%} & \textbf{26.9\%} & \textbf{18.5\%} \\
             \midrule
             
             DSA~\cite{chan2016anticipating}
             & 14.0\% & 4.6\% & 3.8\%    \\
             RRL~\cite{zeng2017agent}
             & 19.0\% & 8.5\% & 4.7\%   \\
             \midrule

             BP~\cite{li2020gcn}
             & 4.2\% & 2.4\% & 1.9\%    \\
             BCP~\cite{li2020make}
             & 6.9\% & 3.9\% & 3.4\%    \\
             
             \midrule
            \bottomrule
        \end{tabular}
        \label{table:consistency}
        \vspace{-3mm}

\end{table}

\subsection{Planning-aware Evaluation}
We study the influence of risk identification on the downstream driving task in Table \ref{table:benchmarkplan}.
Initially, we test the built-in \textit{Auto-pilot} in CARLA to run the collected scenarios. However, \textit{Auto-pilot} can not perform well in \textit{Interactive} and crashes in 58\% of \textit{Obstacle}.
 %
%
We thus adopt the privileged model proposed in~\cite{chen2020lbc} as our planner. 
We compare the planner~\cite{chen2020lbc} using the \textit{Full observation} and restricted observation where the planner only observes the ground truth risk. 
The planner with restricted observation performs slightly inferior in both \textit{Interactive} and \textit{Obstacle} scenarios because we do not want the planner to overfit the recorded trajectories, which will make the evaluation invalid. 

For \textit{Rule-based} algorithms, \textit{Range} achieves favorable results because it owns the nearly full observation by detecting every near object as a risk. 
While \textit{trajectory prediction and collision checking} algorithms perform favorably in terms of \textbf{Recall} in Table ~\ref{table:risk_assessmen_benchmark}, we observe the trend is not aligned in the planning-aware evaluation.
This is because the models cannot successfully predict the risk until the risk is nearby. 
Thus, the planner cannot react in time,
especially for \textit{Obstacle}, requiring more buffer to avoid.
In contrast, while vision-based algorithms are not robust temporally (see PIC results in Table~\ref{table:risk_assessmen_benchmark}), they can predict the risk earlier by incorporating visual information.
Therefore, \textit{Behavior prediction-based} models perform on par with \textit{Trajectory Prediction and Collision Checking} in \textit{interactive}, and better in the \textit{obstacle}. Among them, \textbf{DSA} outperforms all baselines in \textit{interactive} and \textbf{BCP} outperforms all learninig-based methods in \textit{obstacle}. 
%
Combined with the risk localization and anticipation benchmark and the planning-aware evaluation, the results suggest that a risk identification that incorporates visual information is inevitable but requires more effort to improve the robustness, which we will discuss in Section~\ref{sec:discussion}.
\label{subsec:planning-aware-metric}

\section{Discussion}
\label{sec: discussion}

\label{sec:discussion}

\noindent\textbf{Trajectory Prediction and Collision Checking.}
Despite the recent huge progress in trajectory prediction, we find out that the top-ranked method QCNet~\cite{zhou2023query} performs on par with other trajectory prediction methods. 
%
In other words, a high-performing trajectory predictor does not always guarantee improved risk identification performance.
This
is due to shortcomings in the "Collision Checking" mechanism.
Our evaluation, which takes planning into account, further strengthens this observation.
%
%
We hope our findings will encourage the community to collaboratively address the safety challenge from a holistic system perspective.

\noindent\textbf{Temporal Consistency.}
Inspired by the observation in planning-aware evaluation (Section ~\ref{subsec:planning-aware-metric}), we study the temporal consistency of models' predictions, as shown in Table \ref{table:consistency}.
To evaluate temporal consistency, we determine if a risk is predicted accurately and consistently within a specified time frame, leading up to the critical/collision point.
We discover the benchmarked vision-based algorithms lack temporal consistency. 
The finding raises significant reliability concerns for the deployment in the real world. 
Our hypothesis for subpar temporal consistency is that they are trained with per-frame supervision.
We further conduct experiments to demonstrate that using a heuristic smoothing method can significantly improve downstream decision-making in our \textbf{Video Submission}.
This evidence underscores the imperative for more robust spatial-temporal modeling, such as the Transformer~\cite{shi2022motion,vip3d,hu2023_uniad}, to enhance the performance of risk identification and subsequent tasks.

\section{Conclusion}

We introduce RiskBench, 
a scenario-based benchmark focusing on assessing risks induced by dynamic traffic participants and unexpected events. 
We explicitly design a scenario taxonomy for systematic collection of risks and scenario-based performance analysis.
We benchmark four types of existing risk identification algorithms by assessing them using three risk metrics, i.e., risk localization, risk anticipation, and planning awareness. 
%
%
%
In the future, we will explore more advanced risk-aware planners~\cite{chen2021context,diehl2021umbrella,xu2022constraints,diehl2023uncertainty} to enhance robustness of the planning-aware evaluation.


\section*{Acknowledgement}
The work is sponsored in part by the Higher Education Sprout Project of the National Yang Ming Chiao Tung University and Ministry of Education (MOE), the Yushan Fellow Program Administrative Support Grant, and the National Science and Technology Council (NSTC) under grants 110-2222-E-A49-001-MY3, 110-2634-F-002-051, 111-2634-F-002-022-, Mobile Drive Technology Co., Ltd (MobileDrive), and Industrial Technology Research Institute Mechanical and Mechatronics Systems Lab.

\bibliographystyle{IEEEtran}
\bibliography{egbib}

\end{document}